\pdfoutput=1

\documentclass[11pt]{article}

\usepackage[]{ACL}

\usepackage{times}
\usepackage{latexsym}

\usepackage[T1]{fontenc}

\usepackage[utf8]{inputenc}

\usepackage{microtype}
\usepackage{inconsolata}
\usepackage{graphicx}
\usepackage{mathtools}
\usepackage{subfig}
\usepackage{booktabs}
\usepackage{adjustbox}
\usepackage{placeins}

\usepackage{xurl}

\usepackage{amssymb}

\DeclarePairedDelimiter{\round}\lfloor\rceil

\title{Improving Statistical Significance in Human Evaluation \\  of Automatic Metrics via Soft Pairwise Accuracy}

\author{\stepcounter{footnote}Brian Thompson\thanks{{ }{ }Correspondence: \href{brianjt@amazon.com}{brianjt@amazon.com}. Work is unrelated to and conducted independently from the author's position at Amazon.} \\
  Amazon \\
  \\\And
  Nitika Mathur \\
  Oracle \\
  \\\And
  Daniel Deutsch \\
  Google \\
  \\\And
  Huda Khayrallah \\
  Microsoft
  \\
}

\begin{document}
\maketitle
\begin{abstract}
Selecting an automatic metric that best emulates human annotators is often non-trivial, because there is no clear definition of ``best emulates.'' A meta-metric is required to compare the human judgments to the automatic metric scores, and metric rankings depend on the choice of meta-metric. 
We propose Soft Pairwise Accuracy (SPA),
a new meta-metric that builds on Pairwise Accuracy (PA)
but incorporates the statistical significance of both the human judgments and the metric scores.
We show that SPA is more stable than PA with respect to changes in the number of systems/segments used for evaluation. 
We also show that 
PA can only assign a small set of distinct output values to metrics, and this results in many metrics being artificially assigned the exact same PA score. We demonstrate that SPA fixes this issue.
Finally, we show that SPA is more discriminative than PA, producing more statistically significant comparisons between metrics. 
SPA was selected as the official system-level metric for the 2024 WMT Metrics Shared Task.
\end{abstract}

\section{Introduction}

Automatic metrics are crucial because researchers and practitioners in NLP typically can't afford the high cost and latency of high-quality human evaluations. 
Despite their shortcomings, metrics like word error rate and BLEU \cite{papineni-etal-2002-bleu}---in conjunction with carefully curated test sets---have been crucial for the field of NLP, as they have provided a yardstick to make continual progress over many decades in automatic speech recognition and machine translation (MT), respectively.

\begin{figure*}
\centering
\includegraphics[width=.95\textwidth]{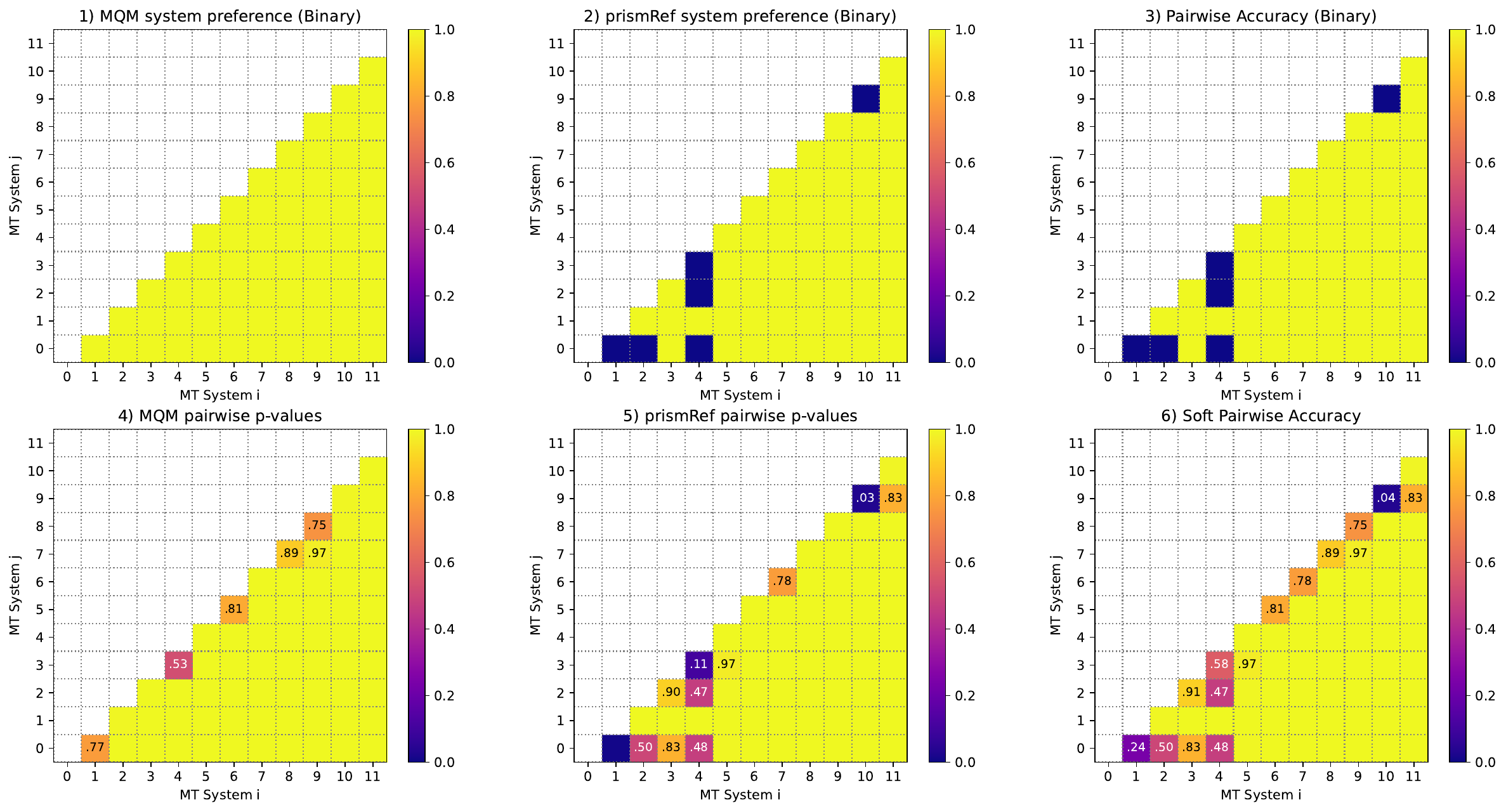}
\caption{Illustration of the individual components used to calculate both SPA and PA for the Prism metric \citep{thompson-post-2020-automatic, thompson-post-2020-paraphrase} 
on the WMT 2023 English-German language pair.
Each box represents a comparison between two systems, systems $i$ and $j$.
MT systems are sorted by average human judgment score for easier interpretation.
The right column is one minus the absolute difference between the human preference for systems $i$ over system $j$ (left column) and the metric preference for system $i$ over system $j$ (middle column).
In PA (top row), human and metric preferences are binarized to 0 and 1, and PA is thus an average of binary terms. In SPA (bottom row), human and metric preferences range from 0 to 1, and as a result SPA is an average of values ranging from 0 to 1. 
SPA can be viewed as a "soft" extension to pairwise accuracy that incorporates both human judgment and metric uncertainty, allowing for partial credit.}
\label{fig:pairwise}
\end{figure*}

Reliance on automatic metrics makes selecting a good automatic metric of paramount importance.
Conceptually, an automatic metric should emulate human judgments. 
Selecting an automatic metric typically entails
generating a set of human judgments for a wide variety of outputs from a large number of different systems,
and selecting the automatic metric that produces scores most similar to the human judgments.
But how do we quantify similarity? To select the 
metric which produces output most similar to human judgements, 
we need a meta-metric to compare metric scores and human judgments. 
Despite nearly two decades of research on MT meta-evaluation, 
the community has not reached a consensus on the choice of a meta-metric. 
Various meta-metrics have been introduced over the years to address problems with prior meta-metrics,
while sometimes creating new problems or re-introducing old ones (see \autoref{sec:history}).

Recent works \cite{mathur-etal-2020-results, kocmi-etal-2021-ship} have argued 
that the primary application of a metric is to choose between two competing systems,
therefore the best metric is the one which produces pairwise system rankings most similar to the pairwise system rankings produced by human judgements.
This led to Pairwise Accuracy (PA) being adopted by the WMT Metrics shared task for the past several years \citep{freitag-etal-2021-results,freitag-etal-2022-results,freitag-etal-2023-results}.
However, this argument omits a key detail: standard best practice when comparing two systems with an automatic metric is to consider not only which system the metric prefers,
but also \emph{whether or not that preference is statistically significant} \cite{koehn-2004-statistical}. 
Thus we argue that metrics should emulate not only the accuracy of human pairwise ranking, but also the confidence or statistical significance of the human pairwise ranking.

To this end, we propose Soft Pairwise Accuracy (SPA),
a new meta-metric which takes into account 
statistical significance of both the metric scores and the human judgments
when evaluating the extent to which the metric in question agrees with the human judgments.
We show that soft pairwise accuracy, as its name implies, can be viewed as a soft (i.e.\ non-binarized) version of PA, and present analysis that demonstrates SPA has several distinct advantages over PA. 
First, we find SPA is more stable with respect to the exact choice of MT systems and segments used.
Second, we show that due to the binarization in its formulation, PA can only assign a small set of distinct output values to metrics, and in practice this results in many metrics being artificially assigned the exact same PA score. We demonstrate that SPA fixes this issue.
Finally, we argue that PA is effectively equivalent to SPA with added noise due to binarization. We show that removing this noise (i.e.\ switching to SPA) results in substantially more statistically significant comparisons between metrics, making SPA a more discriminative and therefore more useful meta-metric. 
Our findings resulted in SPA being selected as the official system-level meta-metric for the 2024 WMT Metrics Shared Task \cite{metrics2024}.

\section{Method}

We propose a simple meta-metric for evaluating automatic metrics given human judgments, which we denote Soft Pairwise Accuracy:

\begin{equation}
    SPA = {{N \choose 2}}^{-1} \,\,
    \sum_{i=0}^{N-1}\sum_{j=i+1}^{N-1} 1 -
\lvert p_{ij}^{h} - p_{ij}^{m} \rvert 
\label{eq:spa}
\end{equation}
where 
$N$ is the number of systems for which we have human judgements and metric scores,
$p_{ij}^{h}$ is the $p$-value for hypothesis that system $i$ is better than system $j$ given the human judgments, 
and 
$p_{ij}^{m}$ is the $p$-value for hypothesis that system $i$ is better than system $j$ given the metric scores. %
The term  ${{N \choose 2}}^{-1} = \frac{2}{N(N{-}1)}$ normalizes the summation by the total number of pairs of systems being compared. 

For each pairwise system comparison, 
we use a permutation test \cite{fisher1935design}
to estimate statistical significance of 
the difference in the means of the segment-level scores from a particular metric (or the human judgements) for the two systems. 
We first randomly split the segment-level scores (ignoring the labels, i.e.\ which MT system produced each segment) into two parts and compute the difference in metric score mean. Repeating this process many times provides a set of mean differences we can reasonably expect under the null hypothesis that the two systems are of the same quality. 
We compute a one-tailed $p$-value by calculating the fraction of the time that the random splits produce differences greater than or equal to the mean difference we observe for the two systems. 

Permutation tests are appealing because they don't require any assumptions about the underlying distribution of the data. This fits our use case well because we cannot assume anything about the distribution of segment-level scores of a metric.\footnote{Metric and human annotation distributions are both highly variable \cite{lo-etal-2023-metric, Knowles_Lo_2024}.}
Permutation tests instead have the assumption of exchangeability \cite{pitman1937significance, draper1993exchangeability, good2002extensions}---that is, 
under the null hypothesis (in our case, that the two MT systems are of equal quality) the joint distribution of the observations is invariant under permutations of the data labels.
To help ensure exchangeability, we perform permutations such that each split has exactly one translation of each test set sentence, commonly referred to as a paired permutation test \cite{good2013permutation}.

Here we present some concrete examples for the sake of intuition.
Suppose a metric reports a $+10$ point difference between system $i$ and system $j$, and that the random permutations only produce a metric difference $\geq 10$ points $1\%$ of the time. Thus $p_{ij}=0.01$ and we conclude that the metric has high confidence that system $i$ is better than system $j$.
Likewise, if the metric reports the systems have a $-10$ point difference, we might find that 
the random permutations produce a metric difference $\geq -10$ points $99\%$ of the time. 
Thus $p_{ij}=0.99$ and we conclude the metric has high confidence that system $i$ is worse than system $j$.
If the systems have the same metric score, we would expect about half of the random permutations to produce a metric difference $\geq 0$ and thus $p_{ij}=0.5$, indicating the metric finds the two systems indistinguishable from each other.

\subsection{Relationship to Pairwise Accuracy}

PA is defined as 

\begin{equation}
    PA = {{N \choose 2}}^{-1} \,\,
    \sum_{i=0}^{N-1}   \sum_{j=i+1}^{N-1} \,a_{ij}^{m}
    \label{eq:pa0}
\end{equation}
where $\,a_{ij}^{m}$ is 1 when the metric scores and human judgments prefer the same system and 0 otherwise. 
PA is equivalent to the Kendall rank correlation coefficient \cite{KENDALL}, modulo a linear scaling and shifting
(see \autoref{kendall}).

A $p$-value $p_{ij}$ will be less than 0.5 when the human raters (or automatic metric) prefer system $i$ over system $j$, and greater than 0.5 when the human raters (or automatic metric) prefer system $j$ over system $i$. 
This allows us to define PA in terms of binarized $p$-values:

\begin{equation}
    PA =  {{N \choose 2}}^{-1} \,\,
    \sum_{i=0}^{N-1}\sum_{j=i+1}^{N-1} 1 -
    \left|  \round{\,p_{ij}^{h}} {-} \round{\,p_{ij}^{m}} \right|
    \label{eq:pa}
\end{equation}

Where binarization is denoted as:
\begin{equation*}
    \round{x} = \begin{dcases*}
         1 & $x \geq 0.5$ \\
         0 & $x < 0.5$ \\
      \end{dcases*}
\end{equation*}

Comparing \autoref{eq:spa} and \autoref{eq:pa} illustrates that SPA can be viewed as a `soft' extension to PA that incorporates uncertainty in both the human and metric scores. A visualization of this is provided in \autoref{fig:pairwise}. 

In cases where both the MT metric and the human evaluation both have high statistical significance (regardless of whether the metric agrees with the human judgments or not, i.e.\ $p_{ij}^{m} \approx 0$ or $p_{ij}^{m} \approx 1$), 
the contribution of that system pair to SPA and PA is approximately identical. 
However, there are two important cases where our meta-metric differs from PA: 
\begin{enumerate} 
    \item The human evaluation has high statistical significance (i.e.\ $p_{ij}^{h} \approx 0$ or $p_{ij}^{h} \approx 1$), but the metric has low statistical significance (i.e.\ $p_{ij}^{m} \approx 0.5$): Even if the metric happens to choose the correct winner, we partially penalize the metric for not having high statistical significance. 
    \item The human evaluation finds the systems are approximately tied (i.e.\ $p_{ij}^{h} \approx 0.5$): In this case, we partially penalize the metric if has high statistical significance 
    (i.e.\ $p_{ij}^{m} \approx 0$ or $p_{ij}^{m} \approx 1$)
    even if it happens to pick the same winner as the human evaluation, and to get full credit the metric must match the human evaluation statistical significance (i.e.\ $p_{ij}^{m} \approx p_{ij}^{m} \approx 0.5$)
\end{enumerate}

\subsection{Addressing Metric Ties in PA}\label{sec:ties_theory}

The fact that PA considers only binary wins/losses (i.e.\ the binarization in \autoref{eq:pa}) results in an interesting shortcoming in PA. 
There are ${{N \choose 2}}$ pairs of $N$ systems, and thus ${{N \choose 2}} + 1$ distinct values that PA can take on ($0/{N \choose 2}$, $1/{N \choose 2}$, ..., ${N \choose 2}/{N \choose 2}$). For example, in WMT 2022 En-De, there are $N=14$ MT systems and thus ${{N \choose 2}}+1 = 92$. 

However, metrics tend to perform better than a random baseline, so only the upper half of the range is actually useful (e.g.\ this leaves 46 distinct values for $N=14$ systems). We find that this results in PA reporting the same scores for several sets of metrics (see \autoref{sec:ties_results}). 
By removing this binarization, SPA has no such issues. 

\section{Experimental Setup}

\subsection{Data}

We conduct experiments on the data from the 2022 and 2023 WMT Metrics Shared Tasks \cite{freitag-etal-2022-results,freitag-etal-2023-results}. In particular, we use the primary language pairs where MQM judgments were collected. We use the MT Metrics Eval V2 toolkit\footnote{\url{https://github.com/google-research/mt-metrics-eval}} to retrieve official shared task scores. 

We make the somewhat arbitrary decision to compare all metrics, including non-primary metrics but excluding QE metrics (i.e.\ reference-free metrics) which provide segment-level scores. 

In order to compute the statistical significance of comparisons between metrics, we make the simplifying assumption that all system-level metrics are the average of their segment-level metric. This is not true for some metrics, including 
BLEU \cite{papineni-etal-2002-bleu} and chrF \cite{popovic-2015-chrf}.
While it would be possible to re-compute BLEU and chrF for each subset, we average the sentence-level versions of these metrics for simplicity.  
To the best of our knowledge, this approach is also taken in recent WMT metrics shared tasks.

\begin{figure*} [!ht]
\captionsetup[subfigure]{labelformat=empty}
\centering
\subfloat[]{\includegraphics[width=0.32\linewidth]{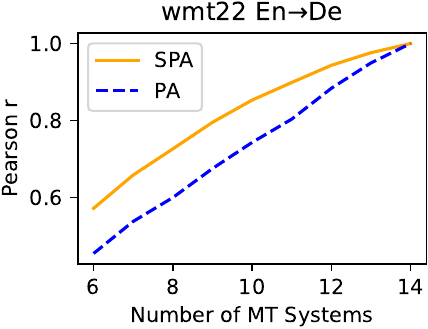}}\hfill%
\subfloat[]{\includegraphics[width=0.32\linewidth]{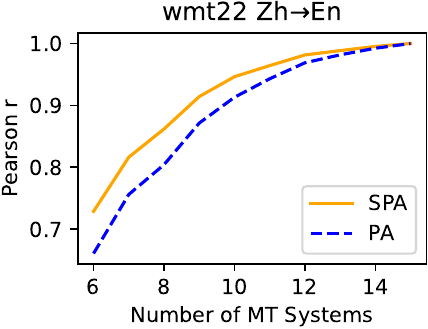}}\hfill%
\subfloat[]{\includegraphics[width=0.32\linewidth]{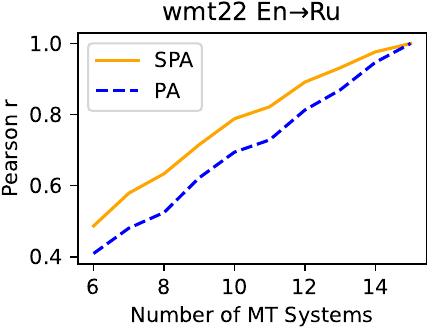}}
\hfil \vspace{-7pt}
\subfloat[]{\includegraphics[width=0.32\linewidth]{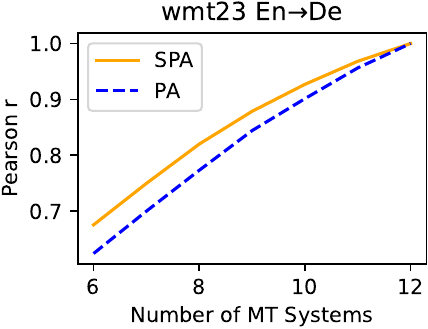}}\hfill%
\subfloat[]{\includegraphics[width=0.32\linewidth]{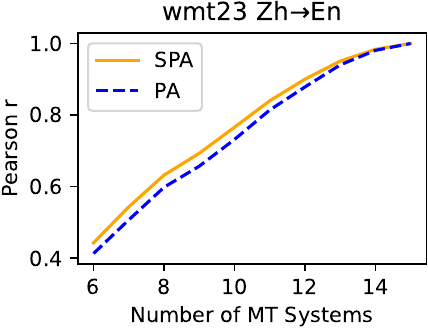}}\hfill%
\subfloat[]{\includegraphics[width=0.32\linewidth]{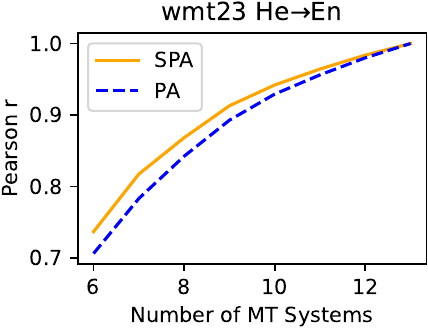}}
 \vspace{-7pt}
\caption{Final metric ranking stability when ablating the number of MT systems (and thus the number of total MQM judgments), measured as change in Pearson correlation coefficient (Pearson $r$) from the ranking computed on all MT systems. Values are averaged over 1000 random trials. We find SPA to be more stable than PA in all cases. }
\label{fig:stability}
\end{figure*}

\subsection{$p$-value Speed Optimization}

We estimate each $p$-value from 1000 random permutations.\footnote{Note that due to the randomness inherent in the $p$-value estimation process, the exact value of SPA can vary slightly from run to run.}
A naive implementation of the paired permutation test is not computationally prohibitive when computing $p$-values for all systems/metrics a single time, but it becomes problematic when we want to compute these values many times in order to estimate statistical significance of metric comparisons.

Experimentally, we find the main speed bottleneck to be generating the random permutations, so when estimating statistical significance of metric comparisons  we cache a batch of permutations and use it for each pair of systems, on a per test-set basis. 
Additionally, by sharing permutations across system pairs, this allows us to pre-compute the 
contribution of each system to means of the random permutations, 
allowing computations to be linear instead of square in the number of systems.
See our code\footnote{\url{https://github.com/thompsonb/mt-metrics-eval/blob/main/mt_metrics_eval/pairwise_paired_permutation_test.py}} for full implementation details.
This results in a speedup of over 1000x compared to the
implementation in Scipy \cite{virtanen2020scipy}.\footnote{Using the `permutation\_test' function from scipy, `permutation\_type' set to `samples' and the `n\_resamples' set to 1000, each $p$-value takes around 40 milliseconds to compute on a laptop.} 
Our speed optimization does not change the computation of a $p$-value for a single system-level comparison, 
but it does mean that the $p$-value for one pair of systems is no longer computed independent from the $p$-value for any other pair of systems. 
Given that we are using these $p$-values as an approximate level of confidence for the system-level comparisons in the SPA meta-metric formulation, as opposed to making any claims about the actual statistical significance of the system-level comparisons, we believe this lack of independence should be inconsequential. 

\begin{figure*}[t]
    \centering
            \includegraphics[width=.97\textwidth]{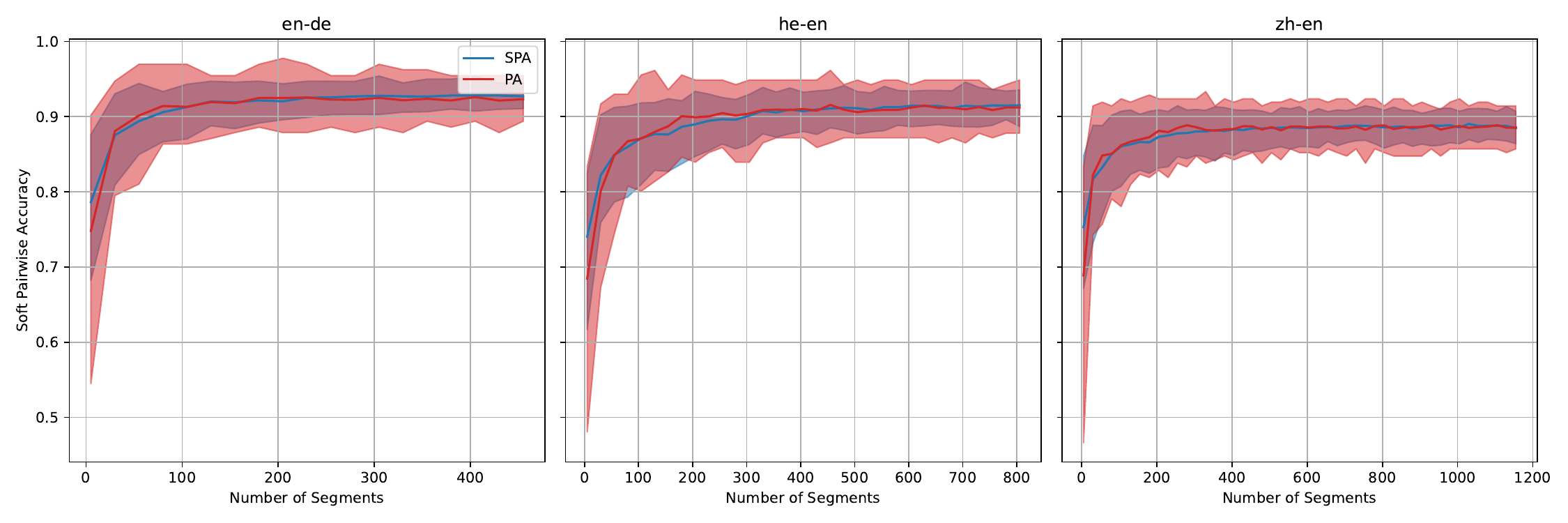}
    \includegraphics[width=\textwidth]{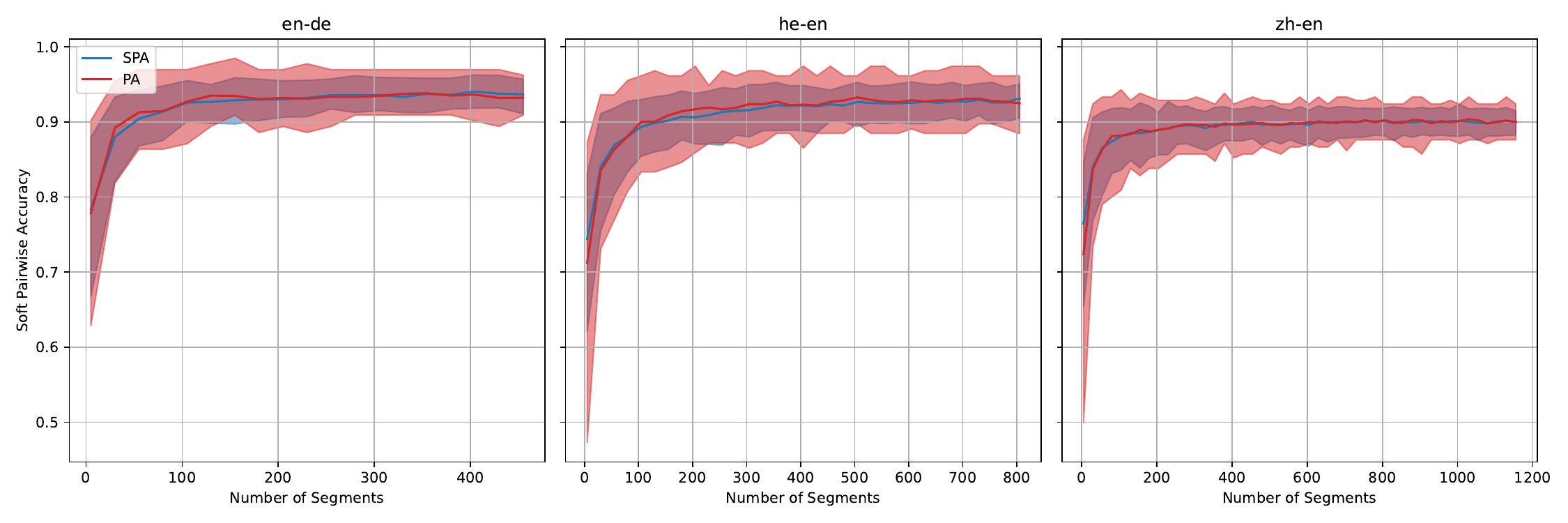}
    \caption{The 95\% confidence intervals for SPA (blue)  and PA (red) on Metric-X (top) and XCOMET (bottom) when varying the number of annotations per system. We find that SPA has a tighter confidence interval, and that the confidence interval shrinks to its full value with smaller sample sizes than PA.}
    \label{fig:sample_size}
\end{figure*}

\section{Analysis}
Meta-metric evaluation is challenging because there is no ground truth (i.e., we don't know the true ranking of the metrics). 
Instead, we conduct analysis to compare SPA and PA.
First, we study how sensitive the meta-metric results are when ablating the number of MT systems and number of segments per MT system, with the assumption that lower sensitivity 
to the exact systems/segments used indicates a better meta-metric.
Second, we examine whether PA indeed has the problem of ties that we hypothesized in \autoref{sec:ties_theory}, and whether SPA fixes this issue.
Finally, we test our hypothesis that the binarization in PA is effectively acting as additive random noise, and that SPA is effectively the same underlying meta-metric with the noise term removed. 

\subsection{Ablation: Number of Systems}

Each year, WMT and the associated metrics task collect and score many online and submitted MT systems.
For an ideal meta-metric, the exact choice of MT systems would have minimal impact on the metric rankings. 
We perform an ablation on the number of MT systems being scored, keeping the number of annotations per system fixed. 
We then compute the correlation (as measured by Pearson's $r$) between the meta-metric's ranking of the ablations  compared to that same meta-metric's full ranking. This allows us to evaluate how sensitive the metric is to the exact selection of MT systems.

When ablating the number of MT systems (and keeping the number of annotations per system fixed), we find (see \autoref{fig:stability}) that SPA is more stable than PA across all MQM language pairs in the last two years of WMT Metrics Shared Tasks.

\subsection{Ablation: Sample Size}
Since SPA relies on the pairwise $p$-values between MT systems, it is also natural to ask how SPA behaves when the number of available segments used for evaluating systems is small since it is harder to find statistical differences between systems with a smaller sample size.
To answer this question, we calculate 95\% confidence intervals for both PA and SPA values of two highly performant metrics---in particular, we considered xCOMET \citep{xcomet} and MetricX-23 \citep{juraska-etal-2023-metricx}---on WMT 2023 using bootstrapping for various numbers of segments, thereby simulating scenarios with less human annotations but a fixed number of MT systems.

\label{results:sample_size}

When ablating the number of segments per MT system (and keeping the number of MT systems fixed), we find (see \autoref{fig:sample_size}) that SPA has tighter 95\% confidence intervals than PA (shown on Metric-X and xCOMET), and that the confidence interval converges to its final value with smaller sample sizes than PA.

\subsection{Ties}\label{sec:ties_results}
As discussed in \autoref{sec:ties_theory}, the binarization in PA limits the number of distinct values it can assign to metrics. 
to ${{N \choose 2}} + 1$. In practice, we find it tends to take on far fewer values. 
For example for WMT 2022 En$\rightarrow$De, PA could theoretically take on 92 distinct values, 
but 
because the metrics fall in a fairly narrow range (PA is $0.626$ for the worst metric and $0.813$ for the best), 
the 21 metrics have only 11 distinct PA scores, with one 5-way PA tie and several 2- and 3-way PA ties (see \autoref{fig:metric_metric_pval_wmt22ende}). 
Since SPA does not binarize each system comparison, it is able to assign any value to each metric, and is therefore potentially better able to distinguish between metrics. 

Results for all language pairs are in \autoref{fig:metric_comp_stats}. We find that on average, PA produces about half as many distinct values as there are metrics while SPA produces one unique value per unique metric.

\subsection{Statistical Significance of Metric Comparisons}

We hypothesize that the binarization in PA is essentially acting as additive random noise on top of the underlying SPA meta-metric. If this is true (and the magnitude of the noise does not dominate the underlying signal), we would expect SPA to produce a similar metric ranking to PA, but with increased statistical significance. 
To test this, we compute statistical significance of the comparisons between each metric
using the PERM-INPUTS \cite{deutsch-etal-2021-statistical} method.
We follow recent shared tasks in greedily computing significance clusters, 
by starting with the highest scoring metric and assigning rank 1 to all metrics until 
we encounter the first metric that is statistically significantly different from \emph{any} previous metric so far.
That metric is assigned rank 2, and the process repeats until all metrics have been assigned a rank.
We echo the shared task organizers' warning that this method can place two metrics that are statistically indistinguishable in different significance clusters (and in the case of PA, we observe this multiple times). 

\label{results:metric_compare_significance}

\begin{figure*}
  \centering
  \begin{tabular}{@{}c@{}}
    \includegraphics[width=.95\linewidth]{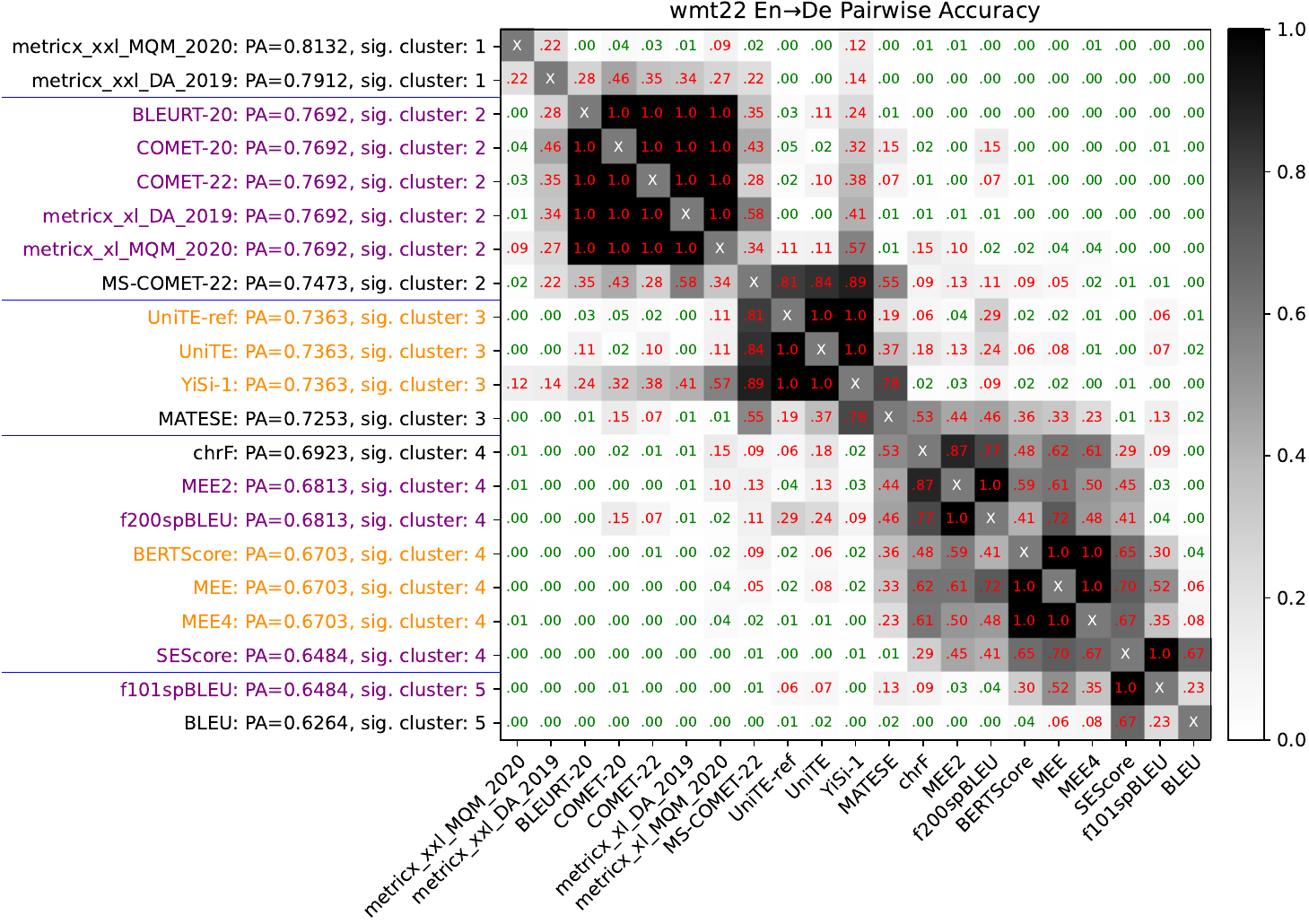} 
  \end{tabular}
  \vspace{\floatsep}
  \begin{tabular}{@{}c@{}}
    \includegraphics[width=.95\linewidth]{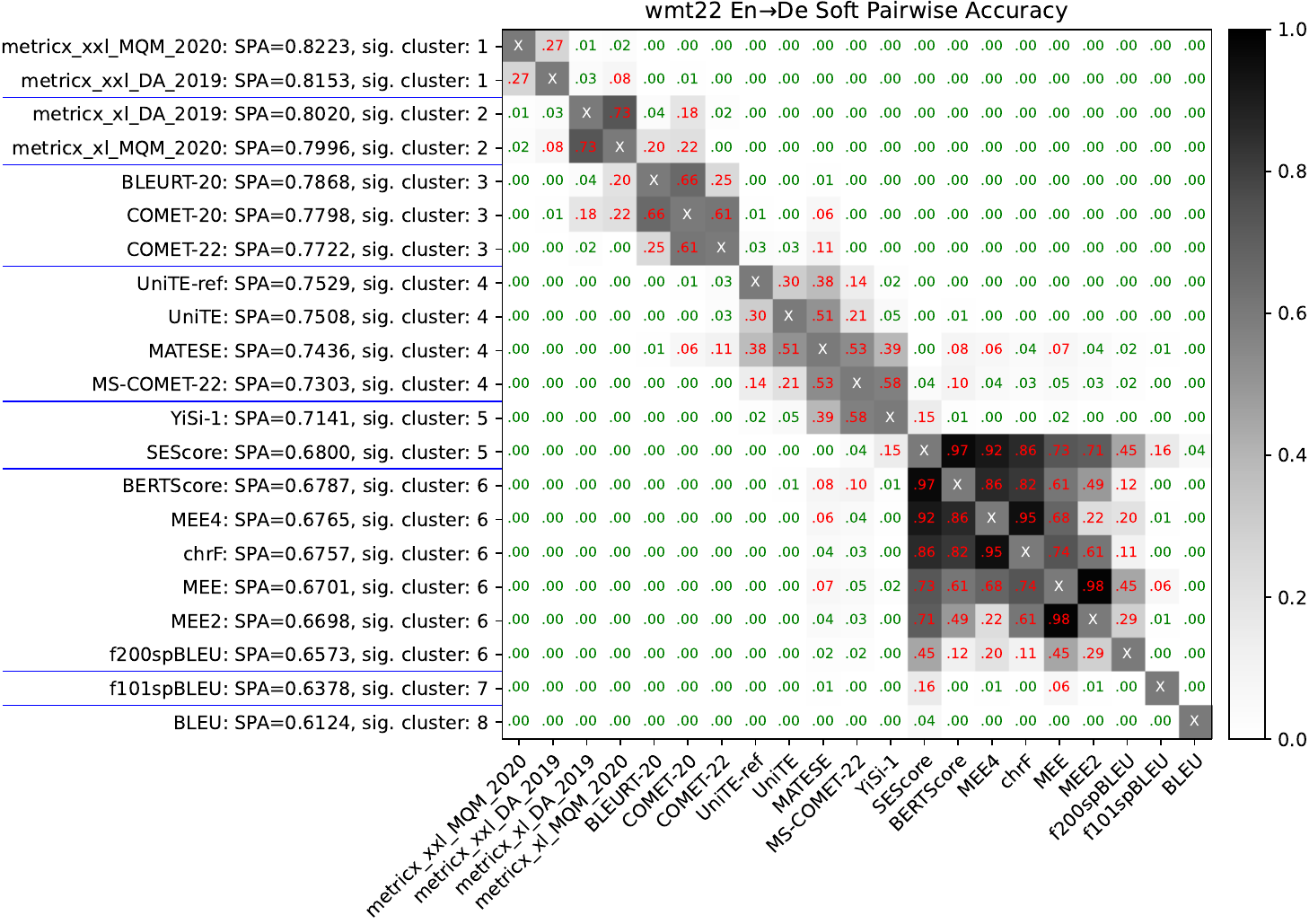} \\[\abovecaptionskip]
  \end{tabular}
  \caption{Metric Comparison Significance, WMT 2022 En$\rightarrow$De.
  Note that PA only assigns 11 distinct values to the 21 metrics (ties are shown in alternating \textcolor[RGB]{128, 1, 128}{Purple} and \textcolor[RGB]{255, 140, 2}{Yellow} text), whereas SPA produces a distinct value for each of the 21 metrics.
SPA produces more statistically significant \textcolor[RGB]{0, 128, 0}{($p$-value <= 0.05, shown in green)} comparisons between metrics (163 vs 108). 
As a result, SPA divides the metrics into 8 significance clusters \textcolor[RGB]{0, 0, 255}{(delineated with blue lines)} compared to only 5 for PA. 
  Results for other language pairs (not shown) are similar. 
  }\label{fig:metric_metric_pval_wmt22ende}
\end{figure*}

On average, SPA increases the number of statistically significant pairwise comparisons by 31\% and the number of significance clusters by 40\% compared to PA, while producing similar scores for each metric (see  \autoref{fig:metric_metric_pval_wmt22ende} for a visualization for WMT 2022 En$\rightarrow$De results and \autoref{fig:metric_comp_stats} for results summary). This is consistent with our hypothesis that PA is effectively SPA with added noise due to binarization.
This means that SPA is a more discriminative, and therefore more useful, meta-metric than PA.

\begin{table*}
\centering
\addtolength{\tabcolsep}{-.5pt}

\begin{tabular}{ccll|lll|lll|lll}

\toprule
        &                 &             &                  &  \multicolumn{3}{|c}{Distinct Metric }                       & \multicolumn{3}{|c}{Significant }              &  \multicolumn{3}{|c}{Significant }           \\
Testset & Language        &  \# MT      &  \# MT           &  \multicolumn{3}{|c}{Values ($\uparrow$)}                    & \multicolumn{3}{|c}{Comparisons ($\uparrow$)}  &  \multicolumn{3}{|c}{Clusters ($\uparrow$)}  \\
        &   Pairs         &  Systems    &  Metrics         &             PA &         SPA              & Max              &  PA & SPA           & Max                      & PA         & SPA         & Max               \\
\midrule
wmt22 & En$\rightarrow$De &    14       &  21              &        11      & \textbf{21}              &   21             & 108 & \textbf{163}  &   210                    &         5  & \textbf{8}  &   21              \\
wmt22 & Zh$\rightarrow$En &    15       &  21              &        12      & \textbf{21}              &   21             & 150 & \textbf{177}  &   210                    &         6  & \textbf{9}  &   21              \\
wmt22 & En$\rightarrow$Ru &    15       &  20              &        10      & \textbf{20}              &   20             & \hspace{1mm} 88 & \textbf{133}  &   190                    &         4  & \textbf{6}  &   20              \\
wmt23 & En$\rightarrow$De &    12       &  25${}^\dagger$  & 12${}^\dagger$ & \textbf{24}${}^\dagger$  & 24${}^\dagger$   & 171 & \textbf{206}  &   276${}^\dagger$        &         5  & \textbf{6}  &  24${}^\dagger$   \\
wmt23 & He$\rightarrow$En &    13       &  25              &        11      & \textbf{25}              &  25              & 180 & \textbf{224}  &   300                    &         5  & \textbf{8}  &   25              \\
wmt23 & Zh$\rightarrow$En &    14       &  25              &        12      & \textbf{25}              &   25             & 186 & \textbf{229}  &   300                    & \textbf{7} & \textbf{7}  &   25              \\
\bottomrule
\end{tabular}
\caption{Number of distinct values produced, number of statistically significant pairwise comparisons (p-value $<=0.05$), and number of statistical significance clusters for PA and SPA.
We provide the best possible value for each category (Max) for comparison, but note that even an ideal meta-metric would likely not achieve this value due to some metrics being highly correlated with each other (e.g. due to training on the same data). 
${}^\dagger$: InstructScore and SEScoreX scores as returned by MT Metrics Eval v2 for WMT23 En-De are identical, causing an exact tie in both PA and SPA. We believe this is an error in MT Metrics Eval v2 but for posterity keep them as-is. 
}\label{fig:metric_comp_stats}
\end{table*}

\section{Historical Context and Related Work}\label{sec:history}

WMT has run a machine translation evaluation since 2006 \cite{koehn-monz-2006-manual}. Since 2007 \cite{callison-burch-etal-2007-meta}, there has also been meta-evaluation of automatic metrics  on the submitted translation systems.
Here we summarize the rich 17 year history of system-level meta-evaluation at the WMT Metrics Shared Tasks\footnote{The WMT Shared Tasks have typically evaluated at both the system- and segment-level, but we focus on system-level meta-evaluation as it is most relevant to our work.} and work related to and directly impacting the shared tasks,
in order to demonstrate how our work fits into the historical context.

In the WMT 2007-2013 metrics evaluations \cite{callison-burch-etal-2007-meta,callison-burch-etal-2008-meta, callison-burch-etal-2009-findings,callison-burch-etal-2010-findings,callison-burch-etal-2011-findings,callison-burch-etal-2012-findings,machacek-bojar-2013-results} Spearman's rank correlation coefficient $\rho$ was used for meta-evaluation of metrics. This was motivated by the fact that Spearman's makes fewer assumptions about the data than the Pearson correlation coefficient. 

The WMT 2013 Translation Shared Task \cite{bojar-etal-2013-findings} introduced system clusters (groups of systems that cannot be distinguished given the human judgments), and the 2013 metrics task \cite{machacek-bojar-2013-results} introduced  empirical confidence of
Spearman's $\rho$ using bootstrap resampling. Since they were not able to resample on the submitted metrics, they only re-sampled human judgments. This iteration also discussed the fact that Spearman's $\rho$ does not give partial credit. The penalty is equal for all wrong judgments, regardless of if  the systems are close or far in quality. To compensate they present additional methods of analysis: Pearson's, and
correlation with systems' clusters from the translation task \cite{bojar-etal-2013-findings}. Those clusters were treated as `ranks with ties,' and then correlation computed against Pearson's and Pearson's correlation against `fuzzy ranks' (the average over ranks of all systems that are not significantly different in human quality).

In 2014, the metrics task \cite{machacek-bojar-2014-results} fully switched to Pearson's $r$  from Spearman's $\rho$. They also did bootstrap resampling to get  empirical confidence intervals of
system level correlations. This change to Pearson's was due to the concerns pointed out in the previous year's shared task, which had explored other meta-metrics. 

The 2015 metrics task \cite{stanojevic-etal-2015-results} continued with Pearson's $r$, and also presented analysis of Pearson's $r$ vs Spearman's $\rho$, and highlighted the instability of Spearman's $\rho$ when MT systems are similar. 

The 2016 metrics task \cite{bojar-etal-2016-results} stuck with Pearson's $r$, but changed the confidence to be the Williams test \cite{williams1959regression}, as \citet{graham-baldwin-2014-testing} had noted that this test is appropriate for dependent correlations.

The 2017 metrics task \cite{bojar-etal-2017-results} kept Pearson's $r$, and Williams test. They also added a pairwise significance test using Williams test. 
This continued in 2018 and 2019 \cite{ma-etal-2018-results, ma-etal-2019-results}

The 2020 metrics task \cite{mathur-etal-2020-results} continued to use Pearson's, but also includes Kendall's Tau for analysis. 
Kendall's Tau is a closer match for the system ranking use case, since it is evaluating whether the ordering of a pair of systems is the same as the human ordering. However, it does not take into account the magnitude difference. 

In 2021, the metrics task \cite{freitag-etal-2021-results} adopted pairwise accuracy \cite{kocmi-etal-2021-ship}, motivated in part by the fact that MT system outliers had an outsized impact on Pearson correlation when it is used to rank MT Metrics \cite{mathur-etal-2020-tangled}. Pairwise accuracy produces the same system-level ranking as Kendall's Tau, as they are equivalent modulo a linear scaling and shifting (see \autoref{kendall}).
The PERM-BOTH hypothesis test of \citet{deutsch-etal-2021-statistical} was used to determine significance. 2021 and 2022 \cite{freitag-etal-2022-results,freitag-etal-2023-results} follow. 

In summary, the historical context of the WMT metric evaluations demonstrates that meta-evaluation is very challenging due to the numerous issues that must be simultaneously addressed, and underscores the pitfalls of making changes to meta-evaluation without considering the full set of ramifications. 
Most relevant to our work, it appears that the switch to pairwise accuracy in 2021 reduced the influence of outliers \cite{mathur-etal-2020-tangled} and (somewhat) aligned meta-evaluation with the standard use of comparing two systems with a metric, but it also reintroduced a problem that was first pointed out by \citet{machacek-bojar-2013-results}
and more fully addressed by the change to Pearson's $r$ from Spearman $\rho$ by
\citet{machacek-bojar-2014-results}: a disregard for the magnitude of differences.
We address this issue by considering empirical confidence, which was first added by \citet{machacek-bojar-2013-results}, and in the process we also better align meta-evaluation to the (more correct) usage of comparing two systems with a metric while also considering the statistical significance of the results.  

\subsection{Relationship to Kendall's Tau}\label{kendall}
Our work builds on pairwise accuracy, 
typically attributed to \citet{kocmi-etal-2021-ship}. 
Pairwise accuracy 
is equivalent to the the widely used Kendall rank correlation coefficient \cite{KENDALL}, modulo a linear scaling and shifting. %
\citet{kocmi-etal-2021-ship} present pairwise accuracy as simply ``accuracy'' and make no mention of its relation to Kendall, which was already in use for MT meta-evaluation \cite{mathur-etal-2020-results}. The term ``pairwise accuracy'' appears to have been coined by \citet{freitag-etal-2021-results} to distinguish it from other types of accuracy.

Kendall's  Tau is defined in terms of concordance (equivalent to our previously defined $a_{ij}^{m}$) and discordance $\,d_{ij}^{m}$, defined to be 1 when the metric and human judgments disagree and 0 otherwise:
\begin{align}
    \tau &= {{N \choose 2}}^{-1} \,\,
    \sum_{i=0}^{N-1}   \sum_{j=i+1}^{N-1} \,(a_{ij}^{m} - d_{ij}^{m} )
    \end{align}
Any system pair which is not concordant is discordant,\footnote{We ignore tie handling, as ties are extremely unlikely in system-level evaluation. Ties in \emph{segment-level} evaluation are an entirely different matter \cite{deutsch-etal-2023-ties}.}
and thus $d_{ij}^{m} =  {1} - a_{ij}^{m}$. Given this and the definition of PA from \autoref{eq:pa0}, we have:
    \begin{align}
    \begin{split}
    \tau  &= {{N \choose 2}}^{-1} \,\,
    \sum_{i=0}^{N-1}   \sum_{j=i+1}^{N-1}  \,  a_{ij}^{m} - (  {1} - a_{ij}^{m})\\
     &= 2 \Bigg({{N \choose 2}}^{-1} \,\,
    \sum_{i=0}^{N-1}   \sum_{j=i+1}^{N-1} \,a_{ij}^{m} \Bigg) -1 \\
    &=  2\; {PA} -1
    \end{split}
    \end{align}

\subsection{Additional Connections to Prior Work}

\citet{graham-liu-2016-achieving} proposed a method of sampling translations from every pair of competing MT systems, creating synthetic systems for scoring. Our work has clear similarities in that we create and score synthetic permutations, but differs in how those synthetic systems are used in the meta-metric formulation. 

\citet{mathur-etal-2020-tangled} showed that MT system outliers had an outsized impact on Pearson correlation. In SPA, outliers impact is limited because $p$-values saturate at 0 or 1. 

\citet{knowles-2021-stability} highlights that as WMT annotation protocols have shifted the original statistical assumptions, and questions the validity of the resulting protocols. Similarly, we show that shifts over the years have caused problems in meta-evaluation. 

\citet{lo-etal-2023-beyond} investigated what magnitude of metric changes tend to be statistically significant. SPA uses statistical significance measures ($p$-values) directly, as opposed to the magnitude of metric differences (e.g.\ as in Pearson correlation). 

\citet{deutsch-etal-2023-ties} demonstrated that principled tie handling is crucial when comparing MT metrics at the segment level, because some metrics produce quantized scores that often result in ties. SPA is system level (i.e.\ sentence level scores averaged over the entire test set), so exact ties are very unlikely. However, SPA can be seen as giving full credit for (statistical) ties, which is similar in spirit.

We show that quantization (specifically binarization) is problematic in PA. Quantization in evaluation has proved problematic in other spaces as well---for example, \citet{schaeffer2024emergent} attributes the widely repeated claim that LLMs have emergent properties to quantization in evaluation.

\section{Conclusions}

We introduce a new meta-metric which we denote soft pairwise accuracy, and 
show that it improves on pairwise accuracy in a number of ways, 
most notably that it is more stable than pairwise accuracy when ablating the number of systems and annotations per system,
it fixes an issue of metric ties observed in pairwise accuracy, 
and it produces more statistically significant comparisons between metrics than pairwise accuracy. We also discuss how soft pairwise accuracy fits into and builds upon the nearly two decade history of meta-evaluation at the WMT Metric Shared Tasks.

\section*{Acknowledgments} 

This work would not have been possible without the WMT shared tasks and metrics tasks releasing their data. We are appreciative of helpful discussions with the WMT metrics task organizers as well as Sweta Agrawal and Rebecca Knowles.

\section*{Limitations}

When computing $p$-values, we assume that system-level metric scores are the average of segment-level metrics scores. There is a line of recent work that seeks to incorporate contextual information into automatic metrics.
Many such works still produce scores at the segment level \citep[e.g.][]{vernikos-etal-2022-embarrassingly, hu-etal-2023-exploring, agrawal2024contexthelpfulchattranslation} but others produce one score per window of a few sentences \cite{raunak-etal-2024-slide} or one score per paragraph \cite{deutsch-etal-2023-training}. 
Our method should still be applicable in such cases, but would require permuting windows or paragraphs instead of segments. 
Additionally, as previously noted, some metrics---notably BLEU \cite{papineni-etal-2002-bleu} and chrF \cite{popovic-2015-chrf}---compute statistics at the segment level and combine them to create document-level scores. Again, permutations would still work but would require some modification. 
To the best of our knowledge, this issue is not limited to our work---the same assumption is made in prior work computing statistical significance of metrics, including the WMT shared tasks \cite{freitag-etal-2021-results, freitag-etal-2022-results,freitag-etal-2023-results} and \citet{deutsch-etal-2021-statistical}.

It is worth noting that the permutations in this work (as in prior works) are done on a single test set, and do not necessarily reflect variations in performance that could result from using the metrics in another domain. Prior work has shown that trained metrics are sensitive to a shift in domain relative to the data domain they were trained on \cite{zouhar-etal-2024-fine}.

\clearpage

\bibliography{main.bbl}

\end{document}